\def\year{2021}\relax
\documentclass[letterpaper]{article} 
\usepackage{aaai21}  
\usepackage{times}  
\usepackage{helvet} 
\usepackage{courier}  
\usepackage[hyphens]{url}  
\usepackage{graphicx} 
\usepackage{natbib}  
\usepackage{caption} 
\usepackage{enumitem}
\usepackage{xr}
\PassOptionsToPackage{dvipsnames}{xcolor}
\usepackage{forest}
\usepackage{pgf}
\usepackage{tikz}
\usetikzlibrary{arrows,automata,shapes.geometric,positioning}
\usepackage{framed}
\usepackage{tikz-qtree}

\usepackage{multirow}
\usepackage{amsthm}
\usepackage{amsmath}
\usepackage{amssymb}
\usepackage{subcaption}
\usepackage{algorithm}
\usepackage{algpseudocode}
\usepackage{listings}
\usepackage{graphicx}
\usepackage[dvipsnames]{xcolor}

\newtheorem{definition}{Definition}

\title{Dynamic probabilistic logic models for effective abstractions in RL}
\author{Harsha Kokel,\textsuperscript{\rm 1} Arjun Manoharan,\textsuperscript{\rm 2}
Sriraam Natarajan,\textsuperscript{\rm 1} \\
Balaraman Ravindran,\textsuperscript{\rm 2} 
Prasad Tadepalli \textsuperscript{\rm 3}\\
}
\affiliations{
\textsuperscript{\rm 1}The University of Texas at Dallas, \\
\textsuperscript{\rm 2}Robert Bosch Centre for Data Science and Artificial Intelligence at
Indian Institute of Technology Madras,  \\
\textsuperscript{\rm 3}Oregon State University\\
\{hkokel,Sriraam.Natarajan\}@utdallas.edu, \{arjunman,ravi\}@cse.iitm.ac.in, tadepall@eecs.oregonstate.edu\\
}

\begin{document}
\maketitle

\begin{abstract}
State abstraction enables sample-efficient learning and  better task transfer in complex reinforcement learning environments.
Recently, we proposed {\sf RePReL}~\cite{KokelMNRT21}, a hierarchical framework that leverages a relational planner to provide useful state abstractions for learning. We present a brief overview of this framework and the use of a {\bf dynamic probabilistic logic model} to design these state abstractions. Our experiments show that {\sf RePReL} not only achieves better performance and efficient learning on the task at hand but also demonstrates better generalization to unseen tasks.

\end{abstract}

\section{Introduction}

Planning and Reinforcement Learning have been two major thrusts of AI aimed at sequential decision making.
We recently presented an integrated architecture we call ``RePReL,''~\cite{KokelMNRT21} which combines relational planning (RP) and reinforcement learning (RL) in a way that exploits their complementary strengths and not only speeds up the convergence compared to a traditional RL solution but also enables effective transfer of the solutions over multiple tasks. In many real world domains, e.g., driving, the state space of offline planning is rather different from the state space of online execution. Planning typically occurs at the level of deciding the route, while online execution needs to take into account dynamic conditions such as locations of other cars and traffic lights. The agent typically does not have access to the dynamic part of the state at the planning time, e.g., future locations of other cars, nor does it have the computational resources to plan an optimal policy in advance that works for all possible traffic events. 

The key principle that enables agents to deal with these informational and computational challenges is {\em abstraction}. In the driving example, the high level state space consists of  coarse locations such as ``O'hare airport''  and high level actions such as take ``Exit 205,'' while the lower level state space consists of a more precise location and velocity of the car and actions such as turning the steering wheel by some amount and applying brakes. Importantly, excepting occasional unforeseen failures, the two levels operate independently of each other and depend on different kinds of information available at different times. This allows the agent to tractably plan at a high level without needing to know the exact state at the time of the execution, and behave appropriately during plan execution by only paying attention to a small dynamic part of the state. 

To achieve the integration of planner and RL effectively, we adapt first-order conditional influence (FOCI) statements \cite{natarajan2008learning} to specify bisimilarity conditions of MDPs \cite{givan2003equivalence}, which in turn help  justify safe and effective abstractions for reinforcement learning \cite{dietterich2000state}. This is the key contribution of this extended abstract -- {\em highlighting the need for an effective StaRAI representation in defining appropriate abstractions for relational MDPs}. We hypothesize and verify empirically that such effective abstractions can result in efficient learning and in many cases, efficient and effective transfer. 

\section{Preliminaries}

We define goal-directed relational MDP (GRMDP) as $\langle S, A, P, R, \gamma, G \rangle$, which is an extension of RMDP definition of \citet{fern2006approximate} for goal-oriented domains by adding a set of goals $G$ that the agent may be asked to achieve. A problem instance for a GRMDP is defined by a pair $\langle s \in S, g \in G \rangle$, where $s$ is a state and $g$ is a goal condition,
both represented using sets of  literals, i.e., positive and/or negative atoms. A solution is a policy that starts from $s$ and ends in a state satisfying $g$ with probability 1. 

The RePReL framework proposes to solve the GRMDPs using a combination of planning and RL in 3 stages:
\begin{enumerate}
    \item {\bf Planning:} Use the hierarchical planner to decompose the goal of the GRMDP to smaller sub-tasks.
    \item {\bf Abstraction:} Get abstract state representations specific to the sub-task.
    \item {\bf RL:} Learn RL agents to perform these sub-tasks in abstract state space.
\end{enumerate}
We defer to \citet{KokelMNRT21} for details of the RePReL framework; here we focus on how to obtain the task-specific abstract state representations using StaRAI approach.

\begin{figure*}[!ht]
    \begin{subfigure}[h]{0.14\textwidth}
    \centering
 \includegraphics[width=0.11\pdfpagewidth, ]{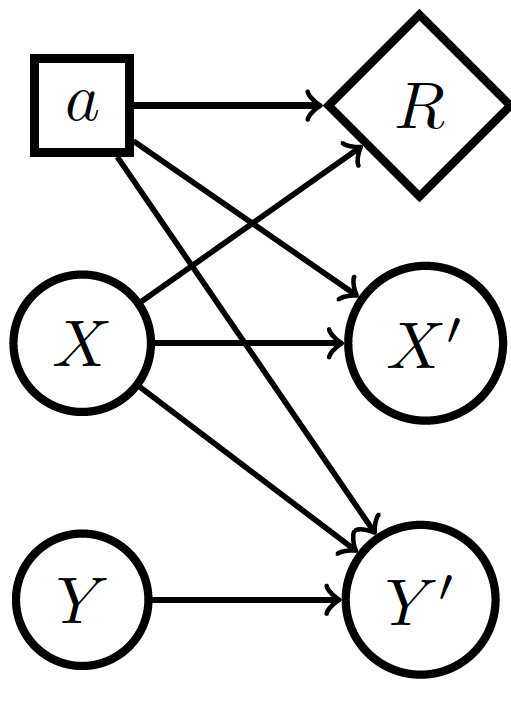}
        \caption{}
        \label{fig:dbn}
    \end{subfigure}
\begin{subfigure}{0.3\textwidth}
\centering
  \fbox{\includegraphics[width=0.22\pdfpagewidth, ]{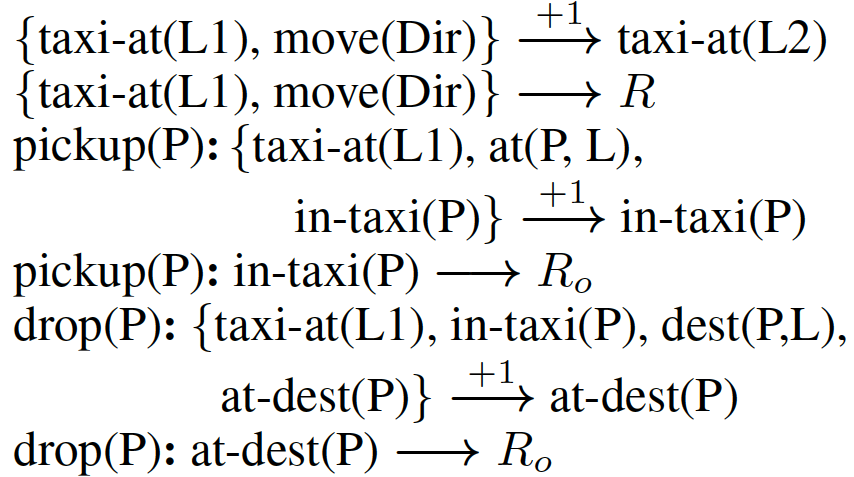}}
\caption{}
\label{fig:taxt_dfoci}
\end{subfigure}
    \begin{subfigure}[h]{0.25\textwidth}
        \centering
        \includegraphics[width=0.21\pdfpagewidth]{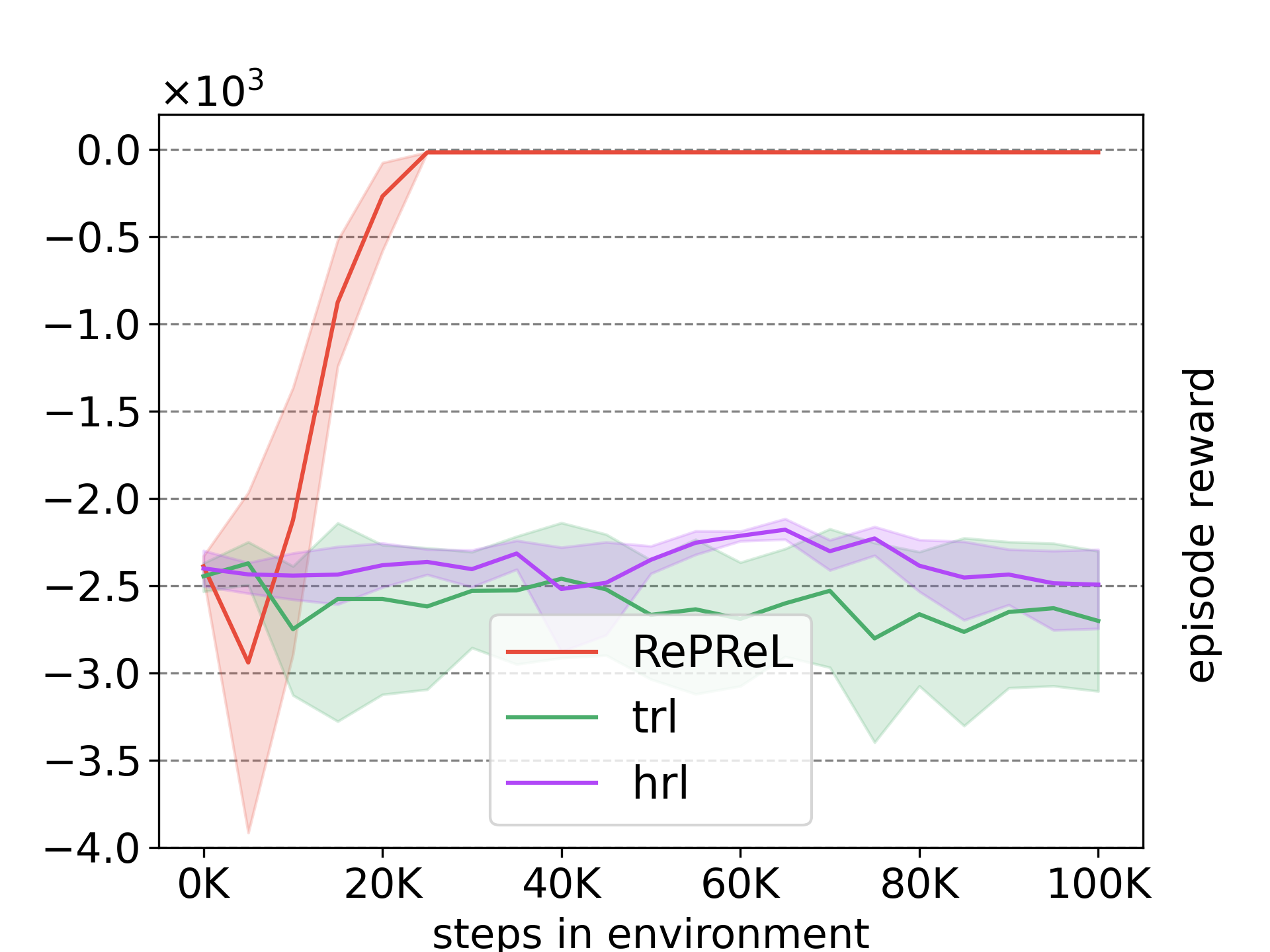}
        \caption{}
        \label{fig:taxi_task1}
    \end{subfigure}
        \begin{subfigure}[h]{0.23\textwidth}
        \centering
        \includegraphics[width=0.21\pdfpagewidth]{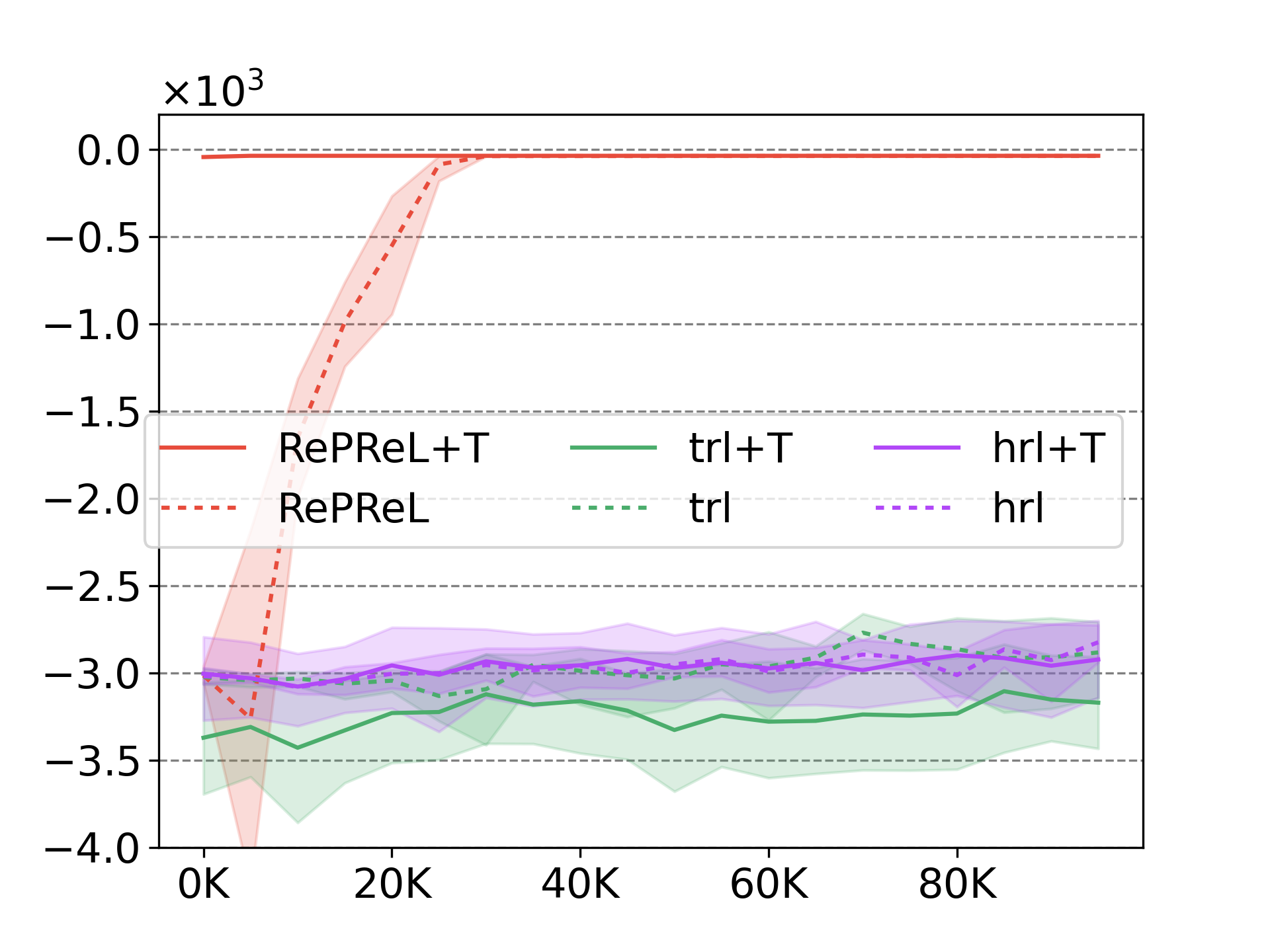}
        \caption{}
        \label{fig:taxi_task2}
    \end{subfigure}
    \caption{\textbf{(a)} A DBN representing influences in an MDP; \textbf{(b)} D-FOCI statements representing influences in a GRMDP of the Taxi domain; \textbf{(c)} Learning curve of RePReL, Taskable RL (trl), and Hierarchical RL (hrl) for task 1, transport $1$ passenger, in the Taxi domain; \textbf{(d)} Learning curve with-transfer (solid lines) and without transfer (dashed lines) for task $2$, transport $2$ passengers, in the Taxi domain. $X$-axis represents steps taken in environments and $Y$-axis represents average episode reward.}
\end{figure*}

\section{Abstraction}

\citet{dietterich2000state} defines {\it irrelevant state variables} for an MDP as variables that never influence the reward function or the relevant state variables. Formally,

\begin{definition} State variables $Y$ are irrelevant in an MDP, if state variables can be partitioned into two disjoint subsets $X$ and $Y$ such that 
\begin{enumerate}
    \item $P(x^{\prime}, y^{\prime}|x, y, a) = P(x^{\prime}|x, a)P(y^{\prime}|x, y, a)$ 
    \item $R(s, a, s^{\prime}) = R(\langle x, y \rangle, a, \langle x^{\prime}, y^{\prime} \rangle) = R(x, a, x^{\prime})$
\end{enumerate}
\end{definition}

Given a Dynamic Bayesian Network (DBN) representation of the transition function of an MDP, like shown in the Fig. \ref{fig:dbn}, the set of irrelevant variables can be identified by starting at the reward variable and collecting all the variables that influence the collected variables.
Note that the {\em transition function of a GRMDP is first-order} and hence a simple DBN does not suffice. Previous work have used a 2-time slice Probablistic Relational Model \cite{guestrin2003generalizing} to capture the transition function in relational MDP. However, they are not suitable for capturing GRMDPs. Hence, we use First-Order Conditional Influence (FOCI) statements~\cite{natarajan2008learning}.
Each FOCI statement is of the form: ``$\text{if } \, \texttt{condition} \, \text{ then } \, {\texttt{X}_1} \, \text{ influence } \, \texttt{X}_2$'', where, \texttt{condition} and $\texttt{X}_1$ are a set of first-order literals and $\texttt{X}_2$ is a single literal. It encodes the information that literal $\texttt{X}_2$ is influenced 
only by the  
literals in $\texttt{X}_1$
when the stated \texttt{condition} is satisfied. 
The influence information captured by the FOCI statements is similar to that captured in many SRL models~\cite{srlbook}.

For GRMDPs, we develop Dynamic FOCI (D-FOCI) statements that express the influence of the literals in the current time step on the literals in the next time-step. Template of the D-FOCI statement is:
\vspace{-0.5em}
$$\texttt{sub-task} \boldsymbol{:}   \texttt{X}_1 \stackrel{+1}{\boldsymbol{\longrightarrow}} \texttt {X}_2$$
\vspace{-1.5em}

\noindent
To distinguish the same time-step and next time-step influences and to capture the temporal influence, we use a $+1$ on the arrow. 
D-FOCI statement encodes the information that for a given \texttt{sub-task}, the literal $\texttt{X}_2$ in the next time-step is directly influenced only by the literals $\texttt{X}_1$ in current time-step. Following the standard DBN representation of MDP, we allow action variables and the reward variables in the two sets of literals. 
To represent
unconditional influences between state literals, we skip the \texttt{sub-task}.

The D-FOCI statements can be viewed as 
2 timeslice PRM and have a similar function of capturing the conditional independence relationships between domain predicates at different time steps, with additional capability of conditioning on sub-task being executed. The Fig. \ref{fig:taxt_dfoci} illustrates the D-FOCI statements in a relational Taxi domain with two sub-task: \texttt{pickup} and \texttt{drop}.
A task-specific model-agnostic abstraction is derived for each sub-task by iterating through the D-FOCI statements and collecting the relevant literals that influence the relevant literals starting with the literals that influence reward variables $R$ and $R_o$. 
The Theorem 1 in \citet{KokelMNRT21} shows that {\em if the GRMDP satisfies the D-FOCI 
statements with any fixed depth unrolling,
then the corresponding 
model-agnostic abstraction has the same optimal value function as the fully instantiated MDP}. We refer to the paper for details.
\section{Evaluation and Discussion}
Our empirical evaluations on $4$ domains show that the proposed task-specific abstraction using the D-FOCI statements have three advantages: {\bf 1.} With abstract state representation, the state space is reduced and hence RePReL achieves better sample efficiency over other methods, {\bf 2.} With task-specific abstractions, RePReL demonstrates efficient transfer across task, {\bf 3.} With the propositionalization for D-FOCI statements, RePReL illustrates zero-shot generalization capability in some cases.
We reproduce the average episode rewards achieved by various agents in the relational Taxi domain on two tasks in Fig. \ref{fig:taxi_task1} and \ref{fig:taxi_task2}. The policies which are transferred from one task to other are indicated with `+T'. As seen, RePReL agents needs significantly less number of samples to reach optimal reward and also show zero-shot generalization across tasks.

A prior work, \citet{nitti2015sample}, introduced sample-based abstraction in hybrid relational domains by regressing the reward and state predicates from the goal state in each sampled episode. Instead, we introduce Dynamic First-Order Conditional Influence statement, an extension of a StaRAI language to capture the conditional influence information of goal-directed relational MDP. We provide an approach for extracting effective task-specific state abstractions in relational MDPs. We demonstrate its utility on learning how to act in relational domains.

\section{Acknowledgements}
\footnotesize
This work is supported by ARO award W911NF2010224 (HK \& SN), AFOSR award FA9550-18-1-0462 (SN), travel grant from RBCDSAI (AM), DARPA contract N66001-17-2-4030 (PT) and NSF grant IIS-1619433 (PT). Any opinions, findings, conclusion or recommendations expressed in this material are those of the authors and do not necessarily reflect the view of the ARO, AFOSR, NSF, DARPA or the US government.

\bibliography{biblio}
\end{document}